\pdfoutput=1

\documentclass[11pt]{article}

\usepackage[]{acl}

\usepackage{times}
\usepackage{latexsym}
\usepackage{graphicx}
\usepackage[T1]{fontenc}

\usepackage[utf8]{inputenc}

\usepackage{microtype}
\usepackage{graphicx}
\usepackage{multirow}
\usepackage{booktabs}
\usepackage{colortbl}

\usepackage{todonotes}

\title{Controlling Topic-Focus Articulation in Meaning-to-Text Generation\\ using Graph Neural Networks}

\author{
Chunliu Wang, Rik van Noord, Johan Bos\\
CLCG, University of Groningen \\
\texttt{\{chunliu.wang, r.i.k.van.noord, johan.bos\}@rug.nl}
}

\begin{document}
\maketitle
\begin{abstract}

A bare meaning representation can be expressed in various way using natural language, depending on how the information is structured on the surface level. We are interested in finding ways to control topic-focus articulation when generating text from meaning. We focus on distinguishing active and passive voice for sentences with transitive verbs. The idea is to add pragmatic information such as topic to the meaning representation, thereby forcing either active or passive voice when given to a natural language generation system.  We use graph neural models because there is no explicit information about word order in a meaning represented by a graph.
We try three different methods for topic-focus articulation employing graph neural models for a meaning-to-text generation task.
We propose a novel encoding strategy about node aggregation in graph neural models, which instead of traditional encoding by aggregating adjacent node information, learns node representations by using depth-first search.
The results show our approach can get competitive performance with state-of-art graph models on general text generation, and lead to significant improvements on the task of active-passive conversion compared to traditional adjacency-based aggregation strategies.
Different types of TFA can have a huge impact on the performance of the graph models.

\end{abstract}

\section{Introduction}

Topic-Focus Articulation (TFA) refers to the way information is packaged within a sentence, in particular the division  of the topic (the given information) and the focus (the new information). 
There are various linguistic devices that determine how topic and focus can be articulated in a sentence: dislocation, word order, active/passive voice, intonation, particles, and more \citep{1984Subject, Lambrecht1994-LAMISA, vall1996, OPUS4-1294, halliday2014introduction}.
TFA plays a crucial role in natural language generation when the input is an abstract representation of meaning and the output is a text. The same information can be realised in various ways, depending on the perspective one takes. Take for example the information provided by the situation where a wolf killed two sheep. We could ask several questions about this situation, viewing it from different points of view:

\begin{quote}
    Q$_1$: What about the sheep? \\
    A$_1$: \textit{The two sheep were killed by a wolf.}\\[10pt]
    Q$_2$: What about the wolf? \\
    A$_2$: \textit{The wolf killed two sheep.}
\end{quote}

A flexible natural language generation system, one that takes as input formal meaning representations, and outputs texts, ideally should be able to package the information related to the perspective being taken.  
In this paper we investigate ways of adding TFA to a formal meaning representation in order to get different texts where information is packaged in different ways. We focus on active-passive alternation in English, as exemplified by the examples above, as in English usually the subject of a sentence is the topic \cite{gundel2004topic}. 
The challenge is to find a simple, intuitive way to add TFA to an abstract meaning representation that gives the desired effects when given to an meaning-to-text NLG component. For this purpose we employ graphical representations of meaning that abstract away from any surface order of the message they convey the meaning of. The NLG modules in our experiments are implemented by neural models. To the best of our knowledge we are the first that experiment with controlling TFA in the context of text generation from formal meaning representations.

The paper is organised as follows.  
In Section~\ref{sec:background} we introduce the semantic formalism of our choice, Discourse Representation Theory \cite{Kamp1993}, and in particular the TFA-neural graph representation of meaning that we use in our experiments. Here we also review previous meaning-to-text approaches for Discourse Representation Structures and graph-based methods for graph-to-text generation task.
In Section~\ref{sec:method} we describe the process of acquiring graph-structured data for graph neural networks (GNNs), and then we focus on presenting the three different TFAs we added in meaning representation.
 The basic idea is to use the simple markers to mark active-voice data and passive-voice data respectively to help the neural models distinguish different types of input graph.
In Section~\ref{sec:experiments} we introduce the implementation settings, and the evaluation metrics.
We present a comparison of local encoders and deep encoders based on three types of GNNs training on different graph representation with three types of TFAs. 



\section{Background}\label{sec:background}

\subsection{Discourse Representation Structures}\label{sec:drs}

Discourse Representation Theory (DRT)
is a well-studied semantic formalism covering many linguistic phenomena including scope of negation and quantifiers, interpretation of pronouns, presupposition, temporal and discourse relations \cite{Kamp1993}. The meaning representation proposed by DRT is the Discourse Representation
Structure (DRS), a recursive first-order logic representation comprising of discourse referents (the entities introduced in the discourse) and relations between them. We work with a particular variant of DRS, namely the one proposed in the Parallel Meaning Bank \cite{abzianidze-parallel}.

There are five types of semantic information that can be found in the PMB-style DRSs: entities, constants, roles, comparison operators, and discourse relations (including negation).
The entities are represented by concepts denoted by WordNet synsets \citep{wordnet}  (for nouns, verbs, adjectives and adverbs), indicating the lemma, part-of-speech and sense number (e.g., \texttt{book.n.02} encodes the entity with the second sense of the noun "book").
The constants are used to represent names, numbers, dates, and deixis. 
The roles (\textit{Theme}, \textit{Agent}, and so on), are represented by the thematic relations proposed in VerbNet \citep{verbnet} and semantically connect two entities.
Comparison operators relate and compare entities or constants.
Discourse relations convey the rhetorical function between different discourse units.

\begin{figure}[!t]
\centering
\includegraphics[scale=.49]{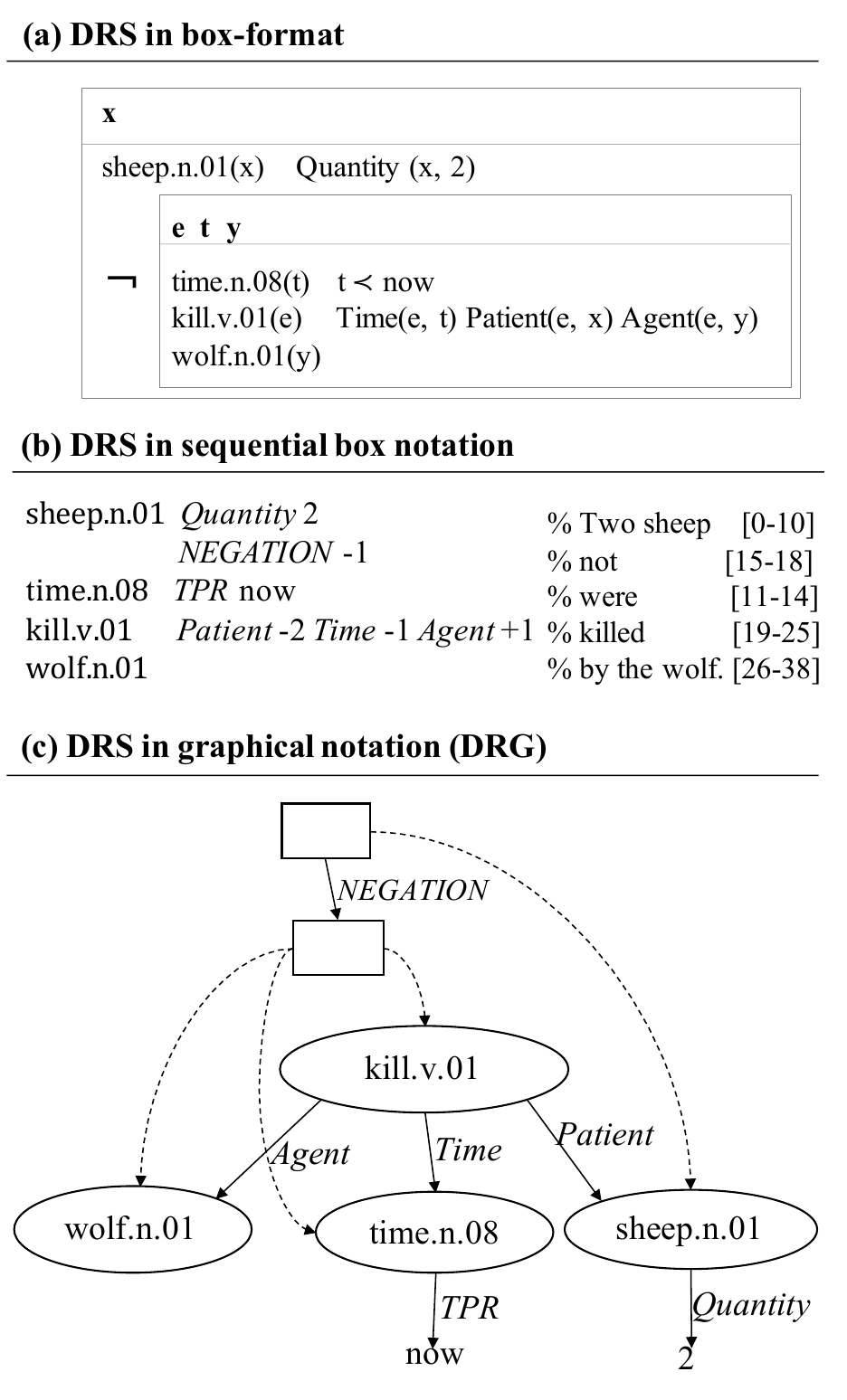}
\caption{(a) Box format DRS for the sentence of \textit{Two sheep were not killed by the wolf}, (b) DRS in sequential box notation, (c) corresponding DRG.}
\label{fig:sbn}
\end{figure}

As Figure~\ref{fig:sbn} shows, there are various ways of representing a DRS. Besides the classic box notation, DRSs can also be displayed in variable-free sequential notation or as directed acyclic graphs, following a recent proposal by \citet{bos2021variable}.
For use in a neural graph model, the DRS data can be converted into a set of triples, using the variable-free sequential box notation. Doing so, the
roles, comparison operators and discourse relations are regarded as edge labels, while entities, constants and discourse units ("boxes") form the nodes in the Discourse Representation Graph (DRG).


\subsection{Generating Text from DRSs}

The purpose of DRS-to-text generation is to produce a text from an input DRS data.
\citet{basile-2015-generation} proposed a pipeline consisting of three components: a surface order module, a lexicalization module that served for constructing an alignment between the abstract structure and the text, and a surface realization module to construct the final output.
\citet{Wang2021EvaluatingTG} uses a standard LSTM model on clause format DRSs to generate text and shows that char-level encoder achieve significant performance,  and \citet{Liu2021TextGF} encode DRS as a tree and uses a sibling treeLSTM model to produce text, which focus on the solution for condition ordering and variable naming in DRS representation.

In this paper, however, as our motivation is to convert freely active and passive voice on DRS data, we consider the graphical representation as input and treat the meaning-to-text generation task as Graph2Seq learning task. A DRG is neutral with respect to word order in the surface realisation of text, and therefore serves our purpose of topic-focus articulation better than a sequential representation of meaning in which order is explicitly encoded.\footnote{A completely different architecture, based on sequential meaning representations, would also be possible, where TFA would take place by operating on the sequential meaning, before given to a Seq2Seq module.}

\subsection{Neural Networks for Graph}

How to encode the input graph is the key issue for the graph-to-text generation task.
A GNN layer computes every node representation by aggregating its neighbor's representation, the design of how to aggregate is what mostly distinguishes the various types of GNNs. 
Graph Convolutional Networks (GCN) learn representations of nodes by summing over the representations of immediate neighborhood of each node \citep{Kipf2017SemiSupervisedCW}, which has been applied to various text generation tasks \citep{damonte2019gen, guo-etal-2019-densely, song-etal-2020-structural} and has achieved remarkable performance.
Some variants use mean or max pooling as aggregation, weighting all neighbors with equal importance, with the same core computation as GCN.
\citet{velickovic2018graph} consider it is unreasonable to assign equal importance to all adjacent nodes of a node and proposed Graph Attention Networks (GAT), which updates each node representation by incorporating the attention mechanism to calculate the importance of adjacent information.
\citet{ggnn} proposed the Gated Graph Neural Networks (GGNN) for the reason that GCN model has difficulty to learn deep layers nodes information, which use a Gated Recurrent Unit (GRU) to facilitate information propagation between local layers.
This model is used in AMR-to-text generation tasks \citep{leonardo-etal-2019-enhancing} and syntax-based machine translation  \citep{beck-etal-2018-graph}.

All the above methods belong to local graph encoding strategies as they are based on \textit{local node aggregation}.
The opposite approach to it is global encoding strategies, which typically based on Transformer and leverage global node aggregation.
We propose deep traversal graph encoders to replace the global graph encoders and local graph encoders. 
Different from the global graph encoder, which compute a node representation based on all nodes in the graph, deep traversal graph encoder focus on capture rich information from the depth-first search of nodes, while avoiding introducing noise from all nodes.

\begin{figure}[!t]
\centering
\includegraphics[scale=.66]{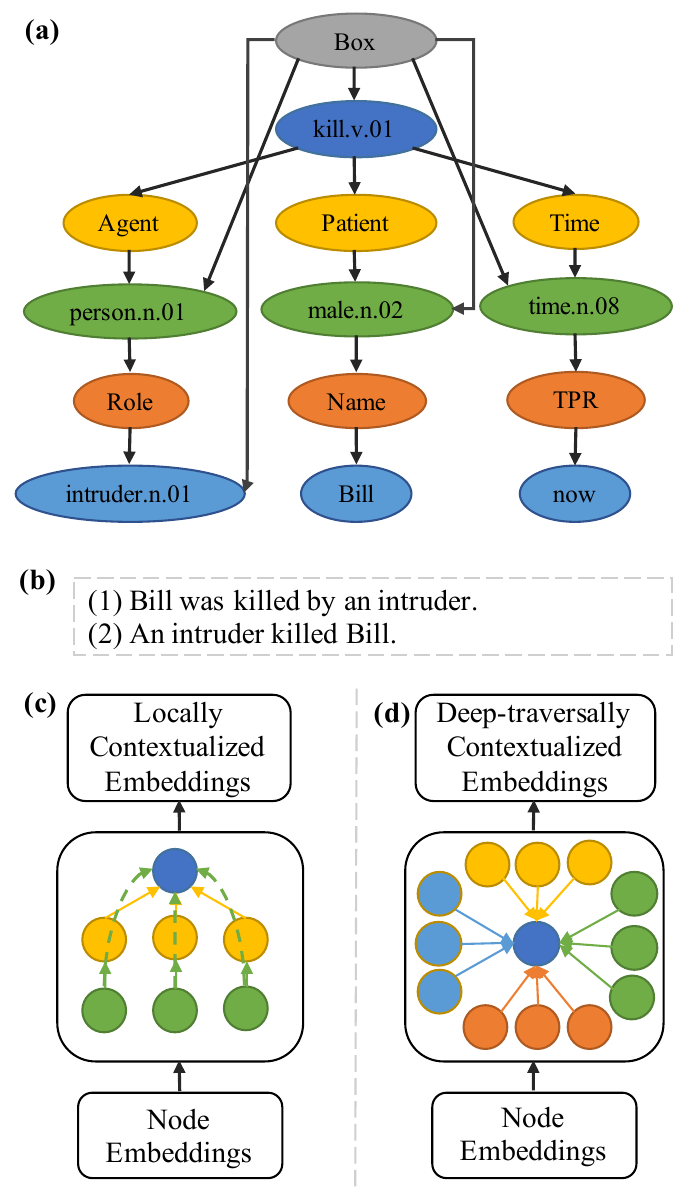}
\caption{(a) Levi graph for a DRG. 
(b) Two possible realisations for (active and passive voice).
(c) Local graph encoder learning information from adjacent nodes,  implicitly capturing 2-hop information by using 2-layers model. 
(d) Deep traversal encoder capturing non-local nodes information by using  depth-first search for a node. }
\label{fig:example}
\end{figure}

\section{Method}\label{sec:method}


\subsection{Graph Preparation}\label{sec:graphprepration}

A DRG is a directed labeled graph defined as $G = (V, E)$, where $V$ is a set of nodes and $E$ is a set of edges.
Each edge in $E$ can be represented as $(v_i, l, v_j)$, 
where $v_i$ and $v_e$ are the indices of incoming nodes and outgoing nodes respectively, and $l$ is the edge label.
We use the extended Levi graph method  \citep{beck-etal-2018-graph} that changes edge labels to additional nodes, and the graph becomes a directed unlabeled graph  where each labeled edge $e = (\textit{$v_i$}, \textit{$l$}, \textit{$v_j$})$ is transformed into two unlabeled edges $e_1$ $=$ (\textit{$v_i$}, \textit{$l$}) and $e_2$ $=$ (\textit{$l$}, \textit{$v_j$}), as shown in Figure~\ref{fig:example}.
This approach is also convenient for representing named entities with more than one token, representing the direction of edges according to the order between tokens of named entities without adding additional edge labels.


\subsection{Adding Topic-Focus Articulation}\label{sec:addingtopic}


A DRG can be textually paraphrased in various ways. For instance, for sentences with a transitive verb, active as well as passive voice could be generated if nothing is known about the information structure. Here we introduce three ways of adding information to the graph about the topic of the sentence (and assume that the rest of the meaning is considered focus).  The idea is that topic controls whether an English sentence with a transitive verb is generated in active or passive voice. We investigate three different methods of adding TFA to a DRG. Our aim is to explore which way of TFA can simply and effectively distinguish active and passive voice, and whether it is convenient for the model to learn this information.
We propose three types of augmenting the DRG with TFA: 

\begin{enumerate} 
    \setlength\itemsep{-1mm}
    \item  Concept $\to$ TOPIC $\to$ Concept (CTC)
    \item  Box $\to$ TOPIC $\to$ Concept (BTC)
    \item  Role $\to$ Role (RTR)
\end{enumerate}

\begin{figure}[!th]
\centering
\includegraphics[scale=.49]{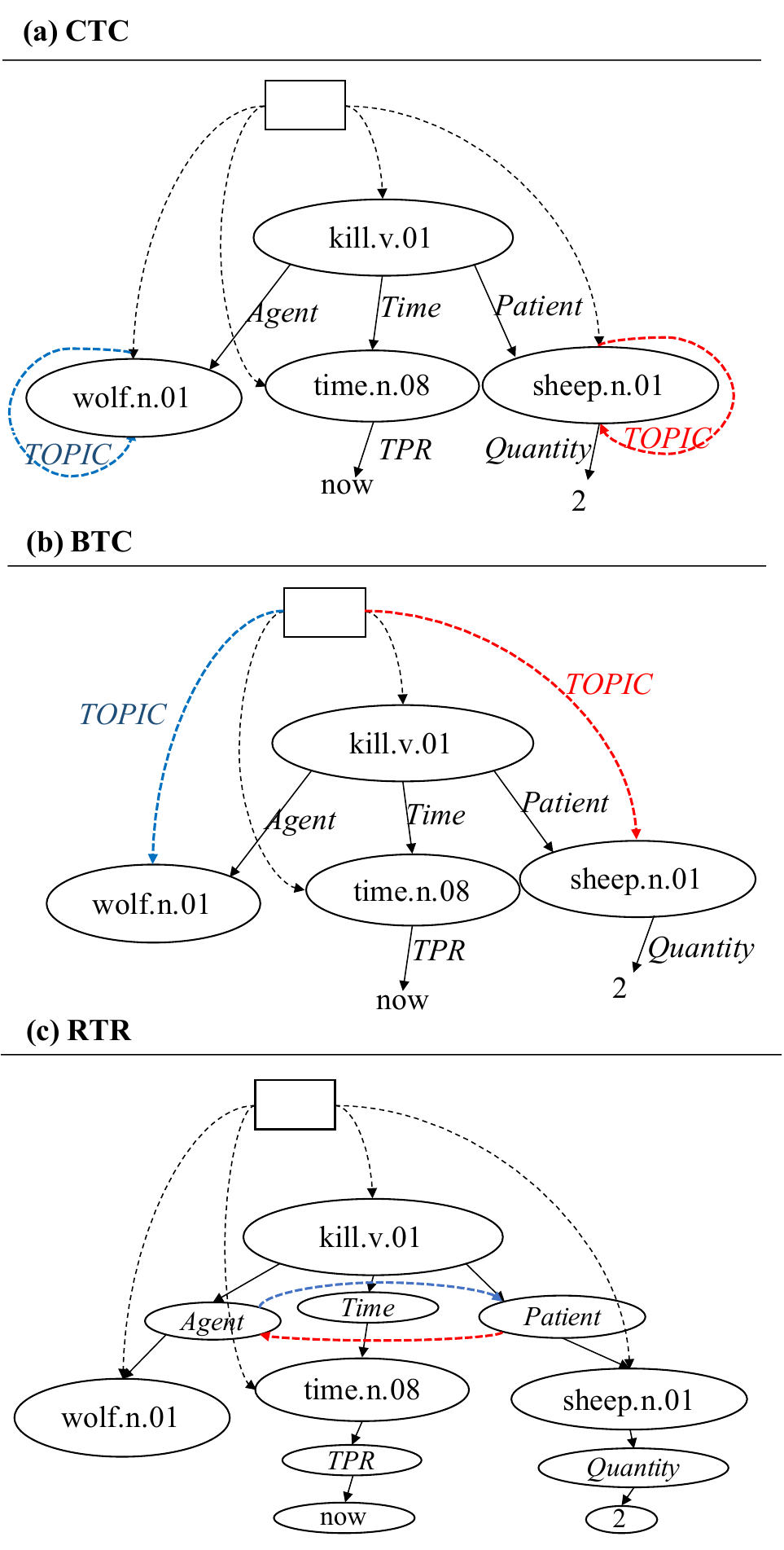}
\caption{Three types of TFA applied to a DRG for \textit{A wolf killed two sheep} (in blue)  or  \textit{The wolf killed two sheep} (in red).}
\label{fig:cases}
\end{figure}

\paragraph{\textit{CTC}} 

Here we specify the subject by giving \textit{TOPIC} marker as edge label and get a self-loop representation.
As shown in Figure~\ref{fig:cases} (a), we give \textit{TOPIC} marker as edge label to the concept \textit{wolf.n.01}, and then \textit{wolf} becomes the subject of text and reference text become active-voice text, vice versa, if \textit{sheep} is the subject, \textit{TOPIC} marker is added as edge label to the concept \textit{sheep.n.01} and the reference text is passive-voice.
This method does not depend on other content in the text, but only focuses on which concept is the subject, and when adding this type TAF in graph and convert graph to Levi graph, the directed acyclic graph will become a directed cycle graph.  

\paragraph{\textit{BTC}}

Here we try to connect more information by giving \textit{TOPIC} as edge label between discourse unit node and subject concept node.
As shown in Figure~\ref{fig:cases} (b), when we use \textit{sheep} as the subject of the reference text, we give \textit{TOPIC} marker as edge label to the concept \textit{sheep.n.01} and its located discourse unit node \textit{Box}, whereas we add \textit{TOPIC} marker between the concept \textit{wolf.n.01} and its discourse node \textit{Box}.
This method considers the global discourse information, and is more intuitive in terms of graph format.

\paragraph{\textit{RTR}}

When we do not know the concept information contained in the data, it is possible to add TFA by using VerbNet role information.
As shown in Figure~\ref{fig:cases} (c), \textit{Agent} is always the VerbNet role connect concept to the verb node, and the text is active-voice if that concept is subject.
\textit{Patient} is one of types of VerbNet roles that connect verb node and the concept of the action or event being imposed by the agent, there are also other types of VerbNet roles introduced in Section~\ref{sec:data}. 
This method do not consider any content of original DRSs data, but by adding the edge such as \textit{Agent}$\to$\textit{Patient} or \textit{Patient}$\to$\textit{Agent} in the graph to decide active-voice and passive-voice.
This representation only based on the transformed extended Levi graph, and essentially changes the flow of information into and out of the graph nodes.


\subsection{Graph Neural Networks}\label{sec:neuralmodel}


Previous research shows it is difficult for local graph encoders to capture non-local nodes information and deeper-layers models perform worse because useful information in earlier layers may get lost \citep{2layer, guo-etal-2019-densely}. 
Inspired by the work of \citet{globallocal}, they propose to integrate GAT and Transformer architectures into a unified global-local graph encoder to learn deep layers nodes information and adopt it on knowledge graphs to text generation task.
In this paper, we propose a deep traversal encoder that aggregates node representations by using all nodes from each node's depth-first search.

Figure~\ref{fig:example} (a) presents a DRG data, the corresponding text is shown in Figure~\ref{fig:example} (b).
For the node \textit{kill.v.01}, it is difficult to access the information from \textit{intruder.n.01} 4-hop away, especially when we use two layers (Figure~\ref{fig:example} (c)), even if we use four layers, the model is very hard to capture.
While when using deep traversal for the node \textit{kill.v.01}, the node \textit{intruder.n.01} has the same distance as node \textit{intruder.n.01}(Figure~\ref{fig:example} (d)), comparing with local graph encoder, deep traversal encoder essentially changes the graph structure, and compared to the global graph encoder, this method preserves more information of related nodes.


\subsubsection{Deep Traversal Encoder}
Compare with the local graph encoder, a deep traversal encoder updates each node information based on depth-first search, which aggregates all the deep traversal nodes instead of adjacency nodes.
In this paper, we compute a layer of the depth-first traversal convolution for a node $i \in V$  based on GGNN, which shows significant performance on the meaning-to-text generation task in previous research \citep{leonardo-etal-2019-enhancing}.



\paragraph{Gated Graph Neural Networks}

The main difference between GCN and GGNN is anologous to the difference between convolutional and recurrent networks \cite{beck-etal-2018-graph}.  GGNN can be seen as a multi-layer GCN where layer-wise parameters are tied and gating mechanisms are added. With this, the model can propagate node information between long distance nodes in the graph \citep{ggnn, beck-etal-2018-graph}.
In particular, the $l$-th layer of a GGNN is calculated as: 

\begin{equation}
\small
h_{i}^{(l+1)} = GRU(h_{i}^{(l)}, \sum_{j\in deep(N_{i})} \frac{W_{dir(j, i)}^{(l)}h_{j}^{(l)}}{|deep(N_{i})|}) 
\label{ggnn}
\end{equation}

where $W_{dir(j, i)}^{(l)}$ represents the weight matrix of the $l$-th layer,   
they are direction-specific parameter, where $dir(j, i)$ $\in$ $\{default, reverse, self\}$, which refer to the original edges, the reversed edges to the original edges, and the self-loop edges \citep{marcheggiani-titov-2017-encoding}.
$deep(N_{i})$ is the the set of nodes by depth-first search for node $i$, the core component of the proposed approach, where the standard GNNs use $N_{i}$, which is the set of immediate neighbors of node $i$. 
$\rho$ is the activation function, we use $ReLU$ in the experiments.
$h_{j}$ is the embedding representation of node $j$ $\in$ $V$ at layer $l$, $GRU$ is a gated recurrent unit, a combination function.



\subsubsection{Decoder}

We adopt standard fully batched attention-based LSTM decoder \citep{bahdanau2016neural}, where the attention memory is the concatenation of the attention vectors among all input words.
The initial state of the decoder is the representation of the output of the encoder. 
In order to alleviate the data sparsity, we add a copy mechanism \citep{gu-etal-2016-incorporating, gulcehre-etal-2016-pointing} on the top of decoder, which favors generating words such as dates, numbers, and named entities that appear in DRGs.


\section{Experiments}\label{sec:experiments}

\subsection{Implementation Details}

\paragraph{Data}\label{sec:data}

We use the data from release 4.0.0 of the Parallel Meaning Bank \citep{abzianidze-parallel}, which contains 10,711 gold instances and 127,302 silver instances.
Each instance contains a text and the corresponding DRS in various formats. We use the sequential box notation to produce the discourse representation graphs (DRGs).
We randomly split the gold data to get 1,000 instances for development and 1,000 instances for test, then we use the remaining gold and silver data as our training data (Table~\ref{table:vocab}).

\begin{table}[htbp]
\centering
\setlength{\tabcolsep}{4pt}
\resizebox{\columnwidth}{!}{
\begin{tabular}{lccccc}
\toprule
& \multicolumn{3}{c}{\textbf{Document-level}} & \multicolumn{2}{c}{\textbf{Word-level}} \\
\hline
\textbf{Data} & \textbf{Train}  & \textbf{dev} & \textbf{test} & \textbf{src} & \textbf{tgt} \\ 
\midrule
\textit{gold}          & 8,711 & 1,000 & 1,000 & 7,856 & 7,699 \\
\textit{gold + silver} & 136,013 &  1,000 & 1,000  & 46,620 & 44,849\\
\bottomrule
\end{tabular}}
\caption{Document statistics and vocabulary sizes.}
\label{table:vocab}
\end{table}

Table~\ref{table:types} shows the distribution
of active and passive voice data as shown in the PMB dataset. 
For convenience, we use VerbNet roles to distinguish the types of active-voice and passive-voice for sentences with transitive verbs in the PMB data.
There are five types of active-voice and passive-voice DRGs representation divided by VerbNet roles. 
The nodes of the VerbNet roles (ingoing or outgoing) determine whether active or passive voice is used.
For example, when the Agent role is an incoming node, this DRG corresponds to passive voice.
For the challenge set of active and passive data, we use all the types of passive voice sentences in the gold data, and the same number of active voice sentences data corresponding different to types (after removing all interrogative sentences, that we disregard for our experiments). 
There are less passive (106) than active sentences (3140) in the PMB data. To get a balanced evaluation set, we draw
randomly 106 from the active instances.

\begin{table*}[!t]
\centering
\setlength{\tabcolsep}{4pt}
\begin{tabular}{l|l|r|r|r|r}
\toprule
\multicolumn{2}{c}{\textbf{ }} & \multicolumn{2}{c}{\textbf{gold data}} & \multicolumn{2}{c}{\textbf{silver data}} \\
\midrule
\textbf{Type} & \textbf{Example of passive} & \textbf{Active} & \textbf{Passive} & \textbf{Active} & \textbf{Passive} \\ 
\midrule
Patient $\to$ Agent      & Bill was killed by an intruder.   & 786 & 43 & 5,285 & 381\\
Theme $\to$ Agent         & The college was founded by Mr Smith. &  1,913 & 42 & 28,120 & 1,125\\
Experiencer $\to$ Agent  & I got stung by a bee.           & 41 & 5 & 311  & 40\\
Result $\to$ Agent        & That was written by Taro Akagawa. & 211 & 13 & 1,241 & 65\\
Source $\to$ Agent       & He was deserted by his friends.   & 189 & 3 & 85 & 32\\
\bottomrule
\end{tabular}
\caption{Distribution of active/passive voice types in gold and silver data in PMB release 4.0.0.}
\label{table:types}
\end{table*}


\paragraph{Setting}

All the models we used implemented based on OpenNMT \citep{klein-etal-2017-opennmt}. 
For the vocabulary, we construct vocabularies from all words, the vocabulary sizes as shown in Table~\ref{table:vocab}.
For the hyperparameters, they are set based on performance on the development set.
We use SGD optimizer with the initial learning rate set to 1 and decay 0.8.
In addition, we set the dropout to 0.5 at the decoder layer to avoid overfitting with batch size 32.
For the deep traversal encoder, we only use one layer for all the graph models.
For the local graph encoder, all the graph models we use two layers, \textit{ReLU} activation and \textit{tanh} highway.
In order to mitigate the effects of random seeds, we report the average for 3 training runs of each model.

\paragraph{Evaluation Metrics}

For automatic evaluation, we use three standard metrics measuring word-overlap between system output and references. They are BLEU \citep{2002bleu},  and METEOR \citep{2007-meteor}, 
which are used as standard in machine translation evaluation and very common in NLG.
%
In order to better evaluate the results of different models, we employ the ROSE manual evaluation metric which proposed by the \citet{Wang2021EvaluatingTG}, 
because this measure is simple to define and easy to reproduce.
That was carried out by creating semantic challenge set and assigning three binary dimensions (either 0 or 1) to each generated text: (1) semantics, (2) grammaticality, and (3) phenomenon.
The result is correct only when the ROSE-score is 1, and we use the ratio of the correct number to the total number of data as the final manual evaluation method accuracy.

\subsection{Automatic Metrics Results}



In order to explore which TFA can better represent the active/passive voice and have a positive impact on the models.
We use automatic metric evaluation methods on the normal test data in order to get the performance for general generation task.
At the same time, we also evaluate the active and passive dataset for obtaining rough performance for this specific tasks.
Table~\ref{table:auto_ggnn_results} is the results of automatic metric results on the normal test and active-passive dataset training on different types of representations by both local graph encoder and deep traversal encoder (see also Appendix~\ref{sec:appendixresults}).

Our results show different types of TFA for the graph-structured data have remarkable different influence for the performance of models on specific tasks (active and passive dataset), but also on general text generation (normal test set).
On the local graph encoder, GGNN has the best performance on both dataset when we use the edge of \textit{RTR} as TFA in the graph.
However, when we use deep traversal encoder, \textit{CTC} become the best TFA, and the normal test performance of the model that using \textit{RTR} as TFA is degraded compared to the local graph encoder, but for active-passive dataset, the deep traversal encoders training with different TFA get the best results compare with the models based on local graph encoder.

\begin{table}[t]
\centering
\setlength{\tabcolsep}{4pt}
\resizebox{\columnwidth}{!}{
\begin{tabular}{ccccc}
\toprule
\multicolumn{1}{c}{\textbf{Model}} & \multicolumn{2}{c}{\textbf{Normal Test}}  & \multicolumn{2}{c}{\textbf{A \& P Test}}  \\
\midrule
\textbf{TFA} & \textbf{BLEU} &\textbf{METEOR}  & \textbf{BLEU} &\textbf{METEOR}  \\
\midrule
\multicolumn{3}{l}{\textbf{Local Graph Encoder}}&\\
RTR & \textbf{55.4} & \textbf{44.0} & \textbf{74.4} & \textbf{54.1}  \\
BTC & 53.8 & 43.3 & 66.9 & 50.4 \\
CTC & 55.0 & 43.6 & 71.8 & 52.6 \\
\midrule
\multicolumn{3}{l}{\textbf{Deep Traversal Encoder}}& \\
RTR & 54.7 & 43.4  & 76.1 & 54.3  \\
BTC & 54.5 & 43.7  & 73.3 & 53.9  \\
CTC & \textbf{55.1} & \textbf{43.8}  & \textbf{75.3} & \textbf{55.7} \\
\bottomrule
\end{tabular}}
\caption{The normal test data results and the active \& passive (A\&P) data results of GGNN models training on DRG representation with different types of TFA.}
\label{table:auto_ggnn_results}
\end{table}

\subsection{Manual Metrics Results}
\begin{table*}[!t]
\centering
\setlength{\tabcolsep}{4pt}
\resizebox{\textwidth}{!}{
\begin{tabular}{l|l|cccc|cccc|c}
\toprule
& &  \multicolumn{4}{c}{\textbf{Passive $\to$ Active}} & \multicolumn{4}{|c}{\textbf{Active $\to$ Passive}} & \multicolumn{1}{|c}{\textbf{ALL}}\\
\midrule
\textbf{Model} & \textbf{TFA}  & \textbf{Sem.} & \textbf{Gram.} & \textbf{Phen.} &\textbf{ROSE} & \textbf{Sem.} & \textbf{Gram.} & \textbf{Phen.} &\textbf{ROSE} &\textbf{ROSE} \\ 
\toprule
\multirow{3}{*}{\textbf{Local Graph Encoder}}&
RTR & 64.2 & 92.5 & 80.2 & \textbf{59.4} & 64.2 & 80.2 & 50.0 & \textbf{34.9} & \textbf{47.2}\\
& BTC & 61.3 & 85.8 & 60.4 & 45.3 & 50.0 & 88.7 & 16.0 & 11.3 & 28.3\\
& CTC & 65.1 & 89.6 & 69.8 & 50.9 & 62.3 & 68.9 & 20.8 & 10.4 & 30.7 \\
\midrule
\multirow{3}{*}{\textbf{Deep Traversal Encoder}}&
RTR &70.8 & 84.9 & 68.9 & 62.3 & 53.8 & 86.8 & 38.7 & 24.5 & 43.4 \\
&BTC & 65.1 & 88.7 & 72.6 & 60.4 & 60.4 & 88.7 & 24.5 & 14.2 & 37.3 \\
&CTC & 73.6 & 93.4 & 87.7 & \textbf{67.0} & 61.3 & 86.8 & 56.6 & \textbf{38.7} & \textbf{52.8} \\
\bottomrule
\end{tabular}}
\caption{Manual metric performance of GGNN models on active and passive challenge set, training on different types of graph data with local graph encoder and deep traversal encoder.}
\label{table:manual_ggnn_results}
\end{table*}

For manual evaluation, in addition to evaluating whether the position of the subject and object is replaced, the transition between active and passive must take into account the grammatical inflection of the verb.
Different types of grammatical inflection of the verb generated by the models will lead to different types of errors, which will affect the three dimensions of manual metrics evaluation at the same time.
In this paper, \textit{phenomena} metric only focus on active and passive voice of verbs in sentences, regardless of semantics and grammaticality.
Due to the limited amount of passive data, we create a challenge set using active and passive data in the gold training data.
All the passive data and active data in DRG can be transformed by changing the TFA, based on this, we can get a challenge set that is different from the data in the training set.
We report the results of accuracy for the active-voice and passive-voice challenge set in Table~\ref{table:manual_ggnn_results}. 

Our results show that for the local graph encoder, the representation by adding \textit{RTR} can get best performance, especially for the task of converting passive data to active data to generate active-voice text, as its \textit{phenomenon} score and \textit{ROSE} are the highest.
When we change the active DRG data to passive DRG data to expect the models to generate passive text, if \textit{CTC} and \textit{BTC} are used as TFA, it is difficult for the models to generate passive voice. They tend to generate active voice text, and almost impossible to generate ideal text, although both semantics and grammaticality are fine.
Compared with local graph encoder, deep traversal encoder can get better performance by using \textit{CTC} and \textit{BTC} as TFA in the DRG,  especially for the \textit{CTC}, the performance for both converting active to passive task and converting passive to active task better than the best results on local graph encoder.

\section{Discussion}\label{sec:analysis}

From the experimental results, for all models based on different TFA and encoders, the difficulty lies in whether the correct passive voice text can be generated.
There are significant gaps of scores between \textit{phenomenon} and the other two dimensions in the task of generate passive voice from Figure~\ref{table:manual_ggnn_results}.
The models intend to generate text in training data with correct semantic and grammar but wrong phenomenon, that means the models cannot learn the TFA very well.
We further subdivide and analyze the types of errors from the \textit{semantic} and \textit{grammaticality} two dimensions (see Appendix~\ref{sec:appendixcase}).


When the edge label in a DRG between the subject and the object connected with the verb is different, which in the Levi graph is the type of the intermediate node for a VerbNet role. 
We find that the local encoder can achieve good results by using \textit{RTR} as TFA. 
This method is somewhat similar to the methods of adding sequential information in a graph used by \citet{beck-etal-2018-graph} and \citet{guo-etal-2019-densely} to improve the performance, but it is equivalent to adding a very small part, so it is not difficult to understand why it achieves better performance.

The deep traversal encoder achieves good results by using the self-loop type TFA (\textit{CTC}). 
The deep traversal encoder essentially obtains all the outflow and inflow nodes information of a node at one time through depth-first search. When we add the \textit{RTR} to the graph, it essentially introduces noise to some nodes in the graph that originally have no direct flow of information, such as the node \textit{Agent} and \textit{sheep.n.01} in Figure~\ref{fig:cases} (e).

However, using the \textit{CTC} in a graph will not have this problem, and through the deep traversal encoder, other nodes in the graph also receive \textit{TOPIC} node information, which is equivalent to increasing the model's memory of \textit{TOPIC} node, so that the models can learn the difference of active and passive in graph structure data.
Intuitively, adding \textit{TOPIC} markers between discourse unit nodes and concepts (\textit{BTC}) can get good performance, but in practice it cannot achieve good performance on both local graph encoder and deep traversal encoder.
We believe that this has a strong correlation with the discourse unit node \textit{Box}, which is very similar to the artificial global node used by \citet{guo-etal-2019-densely} and \citet{caiandlam}, especially when there is only one discourse in the DRG, this node has less influence on the models, and the added edge label \textit{TOPIC} is also difficult to be learned by the models. This has yet to be found where a better way to go to further verification.

\section{Conclusion}

In this paper we propose the use of GNNs to control the generation of active and passive voice text from formal meaning representations.
We use discourse representation structures to represent meaning of sentences, and present the process of how to convert DRS data into graph-structured data for graph models.
On the  graph level, we introduce three types of TFA to distinguish active and passive DRS data.
On the model level, we propose the deep traversal encoder for capturing more information for this task.
We apply one of the most popular graph models, GGNN, to the deep traversal encoder and the local graph model encoder, respectively, to compare the performance of the three TFA.
Our experimental results show that using self-loop labeled TFA (CTC) can achieve optimal results on deep traversal encoder, while using edge-oriented TFA (TRT) can achieve optimal results on local graph encoder.
 Our main contributions can be summarized as follows:

 \begin{enumerate}
     
      \item To the best of our knowledge, we are the first to propose to control TFA in the context of text generation from formal meaning representations; 
    
      \item We present the process of how to convert formal meaning representations  into graph-structured data for graph models;
     
      \item We compare the performance of three graph neural networks, which have not been studied so far for meaning-to-text generation;
      
      \item We introduce deep traversal encoder to replace local graph encoder for capturing more information;
     
      \item We compare three types of TFA in the graph notation of active-voice and passive-voice data to control the models to generate active-passive voice documents.
     
 \end{enumerate}

\bibliography{anthology,custom}
\clearpage

\onecolumn
\appendix

\section{Detailed Results}

\label{sec:appendixresults}

\begin{table*}[h]
\centering
\setlength{\tabcolsep}{4pt}
\begin{tabular}{ll|ccc|ccc}
\toprule
& & \multicolumn{3}{c}{\textbf{Normal Test}}  & \multicolumn{3}{|c}{\textbf{Active \& Passive Test}}  \\
\midrule
\textbf{Model} & \textbf{TFA added} & \textbf{BLEU} &\textbf{METEOR} & \textbf{ROUGE}  & \textbf{BLEU} &\textbf{METEOR} & \textbf{ROUGE}  \\
\midrule
\multicolumn{2}{l}{\textbf{Local Graph Encoder}}&\\
-GCN& RTR& 53.0 & 43.2 & 75.3 & 72.2 & 52.9 & 86.7 \\
& BTC  & 53.4 & 43.1 & 74.9 & 69.5 & 51.6 & 85.3 \\
& CTC& 52.6 & 42.7 & 74.5 & 71.3 & 52.7 & 86.0\\
\midrule
-GGNN& RTR & \textbf{55.4} & 44.0 & 76.2 & \textbf{74.4} & 54.1 & 83.3  \\
& BTC & 53.8 & 43.3 & 75.3 & 66.9 & 50.4 & 83.9\\
& CTC &\textbf{55.0} & 43.6 & 75.7 & \textbf{71.8} & 52.6 & 86.2\\
\midrule
-GAT& RTR & 52.4 & 42.1 & 74.1 & 66.9 & 49.8 & 82.8 \\
& BTC  & 51.9 & 41.9 & 73.7 & 63.7 & 48.0 & 81.3 \\
& CTC & 51.8 & 42.1 & 73.9 & 62.0 & 47.5 & 81.2\\
\midrule
\multicolumn{2}{l}{\textbf{Deep Traversal Encoder}}& \\
-GCN& RTR& 52.4 & 42.6 & 74.3 &  74.0 & 53.8 & 86.5  \\
& BTC  & 53.1 & 43.1 & 74.5 &  72.1 & 52.7 & 85.9 \\
& CTC & 53.8 & 43.0 & 74.9 & 74.4 & 54.3 & 87.5\\ 
\midrule
-GGNN& RTR & \textbf{54.7} & 43.4 & 75.4 & \textbf{76.1} & 54.3 & 87.6 \\
& BTC  & 54.5 & 43.7 & 75.9 & 73.3 & 53.9 & 86.9 \\
& CTC & \textbf{55.1} & 43.8 & 75.7 & \textbf{75.3} & 55.7 & 87.9 \\
\midrule
-GAT& RTR & 50.9 & 41.5 & 72.9  & 65.3 & 48.8 & 82.2 \\
& BTC & 51.6 & 41.8 & 73.4  & 66.7 & 48.7 & 82.1 \\
& CTC & 52.1 & 41.9 & 73.8 & 68.3 & 50.5 & 83.6 \\
\bottomrule
\end{tabular}
\caption{The test data results and the active-voice \& passive-voice data results on different GNN models with local graph encoder and deep traversal encoder training on silver and gold data.}
\label{table:auto_normal_results}
\end{table*}
\clearpage

\section{Fine-grained Error Analysis}
\label{sec:appendixcase}

\paragraph{Semantic} 

Given a DRG, when the models can correctly generate the word order of subject/object and correct grammaticality, but cannot generate the correct verb format for active and passive voice, it will change the agent of sentences, we call this error type \texttt{wrong agent}.
This type of error occurs abundantly in the data representation method of \textit{BTC} in local graph encoders.
Although the deep traversal encoder models has improved, it is still not ideal compared to other representation methods.
One type of the error is about generating repeated subject or object.
Maybe generated text has correct grammar and phenomenon, but the object and subject are repeated, it changed the semantic of original data, we call this error type is \texttt{repeated O/S}.
Some semantic errors are focused on \texttt{named entities}.
When the name entities have more than one token, especially there are more than three tokens in name entities, it would be difficult for both encoders to generate correct name entities. 
We believe this errors can be omitted by using anonymous methods \citep{konstas-etal-2017-neural}, while it requires alignment rules and post-processing work.
In addition, some generated text by models will miss information and lose original semantic of references, we call it \texttt{misinformation}. 
If the models generate the wrong concept and change original semantic of references, we call it \texttt{wrong concept}.
Another common error in local graph encoders is that the \texttt{adjectives} in the subject and object of the generated sentence are crossed, so they are in the wrong place. Deep encoders can handle this problem very well.
See Table~\ref{table:manual_results1} and Table~\ref{table:manual_results2}.

\paragraph{Grammaticality}
Grammatical errors are mainly concentrated in the \texttt{verb format}, especially for passive voice to generate past auxiliary verb.
For example,  \textit{is eaten} corresponds to the active verb \textit{eat}, but the two verb tenses are actually different. \textit{is eaten} consists of the auxiliary verb to be in its present simple form plus the past auxiliary form of the verb to \textit{eat}.
Sometimes, even if the model can successfully generate the subject and object in the correct position, it is difficult to generate the verb in the correct format, especially the verb form does not exist in the training corpus.
Some grammaticality errors are about \texttt{polarity}, this error is common to all models, especially for nouns with low word frequency in the training corpus.
In addition, because the concept is represented by \texttt{WordNet}, which different from AMR, the model using the copy mechanism will directly copy the WordNet after encountering a wordnet with a low word frequency, thereby generating an incorrect sentence.
All the above errors are showned in Table~\ref{table:manual_results2}.

\begin{table*}[!h]
\centering
\setlength{\tabcolsep}{4pt}
\begin{tabular}{l|l|cccc}
\toprule
\textbf{Model \& Error type} &  \textbf{Generated Examples} & \textbf{Sem.} & \textbf{Gram.} & \textbf{Phen.}\\ 
\midrule
\multicolumn{4}{l}{\textbf{Semantic: Wrong name entity}}   \\
\multicolumn{4}{l}{\textbf{Correction: Metin Kaplan murdered Ibrahim Sofu.}} &  \\
\hline
Local Graph Encoder: RTR & Ibrahim Sofu murdered \textcolor{red}{Ibrahim Kaplan}. &  0 & 1 & 1\\
Local Graph Encoder: BTC & Ibrahim Sofu was murdered \textcolor{red}{Ibrahim Kaplan}. &  0 & 0 & 0 \\
Local Graph Encoder: CTC & Ibrahim Sofu was murdered Metin Kaplan. & 1 & 0 & 0 \\
Deep Traversal Encoder: RTR & Metin Kaplan murdered Ibrahim Sofu. &  1 & 1 & 1\\
Deep Traversal Encoder: BTC & Metin Kaplan murdered Ibrahim Sofu.& 1 & 1 & 1 \\
Deep Traversal Encoder: CTC  & Metin Kaplan murdered \textcolor{red}{Ibrahim}. & 0 & 1 & 1 \\
\midrule
\multicolumn{4}{l}{\textbf{Semantic: Wrong concept}}   \\
\multicolumn{4}{l}{\textbf{Correction:The giant hornet killed Tom.}} &  \\
\hline
Local Graph Encoder: RTR & The \textcolor{red}{bank} killed Tom. &  0 & 1 & 1\\
Local Graph Encoder: BTC & Has Tom killed Tom? &  0 & 1 & 0 \\
Local Graph Encoder: CTC & The Tom was killed by Tom. & 0 & 1 & 1 \\
Deep Traversal Encoder:  RTR &  The giant was killed Tom. &  0 & 0 & 0\\
Deep Traversal Encoder: BTC & The giant\_hornet.n.01 killed Tom. & 1 & 0 & 1 \\
Deep Traversal Encoder: CTC   & The \textcolor{red}{tsunami} killed Tom. & 0 & 1 & 1 \\
\midrule
\multicolumn{4}{l}{\textbf{Semantic: Wrong adjective}}   \\
\multicolumn{4}{l}{\textbf{Correction: A old man chased a young girl.}} &  \\
\hline
Local Graph Encoder: RTR & The old man chased the \textcolor{red}{old} girl. &  0 & 1 & 1\\
Local Graph Encoder: BTC & The \textcolor{red}{young} man chased the \textcolor{red}{old} girl. &  0 & 1 & 1 \\
Local Graph Encoder: CTC & The \textcolor{red}{young} man chased the \textcolor{red}{old} girl. & 0 & 1 & 1 \\
Deep Traversal Encoder: RTR & The old man chased a young girl. &  1 & 1 & 1\\
Deep Traversal Encoder: BTC  & The old man chased a young girl. & 1 & 1 & 1 \\
Deep Traversal Encoder: CTC   & The old man chased the young girl. & 1 & 1 & 1 \\
\midrule
\multicolumn{4}{l}{\textbf{Semantic: Misinformation}}   \\
\multicolumn{4}{l}{\textbf{Correction: Schubert composed the famous song "Ave Maria".}} &  \\
\hline
Local Graph Encoder: RTR & Schubert was composed \textcolor{red}{a famous song}. &  0 & 0 & 0\\
Local Graph Encoder: BTC &	Schubert was composed \textcolor{red}{a famous song}. &  0 & 0 & 0 \\
Local Graph Encoder: CTC & Schubert was composed to \textcolor{red}{a famous song}.& 0 & 0 & 0\\
Deep Traversal Encoder: RTR & Schubert was composed \textcolor{red}{"Ave Maria"}. &  0 & 0 & 0 \\
Deep Traversal Encoder: BTC & Schubert composed the famous song "Ave Maria".& 1 & 1 & 1 \\
Deep Traversal Encoder: CTC  & Schubert composed \textcolor{red}{famous songs}. & 0 & 1 & 1 \\
\midrule
\multicolumn{4}{l}{\textbf{Semantic: Wrong agent}}   \\
\multicolumn{4}{l}{\textbf{Correction: The bee stung me.}} &  \\
\hline
Local Graph Encoder: RTR & The bee was stung by me. &  0 & 1 & 0\\
Local Graph Encoder: BTC & The bee was stung by me. &  0 & 1 & 0\\
Local Graph Encoder: CTC & The bee was stung by me. & 0 & 1 & 0 \\
Deep Traversal Encoder: RTR & The bee stung me. &  1 & 1 & 1\\
Deep Traversal Encoder: BTC & The bee was stung by me. & 0 & 1 & 0 \\
Deep Traversal Encoder: CTC   & The bee stung me. & 1 & 1 & 1 \\
\bottomrule
\end{tabular}
\caption{The sample examples according to its types of errors generated by models.}
\label{table:manual_results1}
\end{table*}

\begin{table*}[!h]
\centering
\footnotesize
\setlength{\tabcolsep}{4pt}
\resizebox{\textwidth}{!}{
\begin{tabular}{p{5cm}|p{5cm}|cccc}
\toprule
\textbf{Model \& Error type} &  \textbf{Generated Examples} & \textbf{Sem.} & \textbf{Gram.} & \textbf{Phen.}\\ 
\midrule
\multicolumn{4}{l}{\textbf{Semantic: Repeated O/S}}   \\
\multicolumn{4}{l}{\textbf{Correction: An intruder killed Bill.}} &  \\
\hline
Local Graph Encoder: RTR & \textcolor{red}{Bill} killed Bill. &  0 & 1 & 1\\
Local Graph Encoder: BTC & \textcolor{red}{Bill} killed Bill. &  0 & 1 & 1 \\
Local Graph Encoder: CTC & \textcolor{red}{Bill} killed Bill. & 0 & 1 & 1 \\
Deep Traversal Encoder: RTR & The intruder killed Bill. &  1 & 1 & 1\\
Deep Traversal Encoder: BTC & The intruder killed Bill. & 1 & 1 & 1 \\
Deep Traversal Encoder: CTC  & The intruder killed Bill. & 1 & 1 & 1 \\
\midrule
\multicolumn{4}{l}{\textbf{Semantic: Addition}}   \\
\multicolumn{4}{l}{\textbf{Correction: The dishes were rinsed by her.}} &  \\
\hline
Local Graph Encoder: RTR & She rinsed the dishes \textcolor{red}{from the dishes}. &  0 & 1 & 0\\
Local Graph Encoder: BTC & The dishes was rinsed by herself. &  1 & 0 & 1 \\
Local Graph Encoder: CTC & The dishes rinsed her.& 0 & 1 & 0 \\
Deep Traversal Encoder: RTR & She rinsed the dishes.&  1 & 1 & 0\\
Deep Traversal Encoder: BTC & She rinsed her from the dishes. & 0 & 1 & 0 \\
Deep Traversal Encoder: CTC & She rinsed her out of the dishes.& 0 & 1 & 0 \\
\midrule
\multicolumn{4}{l}{\textbf{Grammaticality: Verb format}}   \\
\multicolumn{4}{l}{\textbf{Correction: The big fish was captured by the old man.}} &  \\
\hline
Local Graph Encoder: RTR  & The old fish was captured by a big man. & 0 & 1 & 1\\
Local Graph Encoder: BTC & The old fish captured the big man. & 0 & 1 & 0\\
Local Graph Encoder: CTC &The old man \textcolor{red}{captured by} the old man. &  0 & 0 & 0  \\
Deep Traversal Encoder: RTR & The big fish captured the man. &  0 & 1 & 0 \\
Deep Traversal Encoder: BTC & The big fish captured an old man. & 0 & 1 & 0 \\
Deep Traversal Encoder: CTC & The big fish \textcolor{red}{captured by} the old man. &  1 & 0 & 0 \\
\midrule
\multicolumn{4}{l}{\textbf{Grammaticality: Polarity}}   \\
\multicolumn{4}{l}{\textbf{Correction: The ten eggs were boiled by my mother.}} &  \\
\hline
Local Graph Encoder: RTR & My mother was boiled by ten eggs. &  0 & 1 & 1\\
Local Graph Encoder: BTC & My mother boiled ten eggs. &  1 & 1 & 0 \\
Local Graph Encoder: CTC &	My mother boiled by ten eggs. & 0 & 0 & 0 \\
Deep Traversal Encoder: RTR & The ten eggs \textcolor{red}{was} boiled by my mother. &  1 & 0& 1\\
Deep Traversal Encoder: BTC & Ten eggs were boiling by my mother. & 1 & 0 & 0 \\
Deep Traversal Encoder: CTC & My mother was boiling by my mother. & 0 & 0 & 0 \\
\midrule
\multicolumn{4}{l}{\textbf{Grammaticality: WordNet}}   \\
\multicolumn{4}{l}{\textbf{Correction: The giant hornet killed Tom. }} \\
\hline
Local Graph Encoder: RTR & The bank killed Tom. &  0 & 1 & 1\\
Local Graph Encoder: BTC & Has Tom killed Tom? &  0 & 1 & 0 \\
Local Graph Encoder: CTC & The Tom was killed by Tom. & 0 & 1 & 1 \\
Deep Traversal Encoder: RTR &  The giant was killed Tom. &  0 & 0 & 0\\
Deep Traversal Encoder: BTC & The \textcolor{red}{giant\_hornet.n.01} killed Tom. & 1 & 0 & 1 \\
Deep Traversal Encoder: CTC  & The tsunami killed Tom. & 0 & 1 & 1 \\
\bottomrule
\end{tabular}}
\caption{The sample examples according to its types of errors generated by models.}
\label{table:manual_results2}
\end{table*}

\end{document}